\pdfoutput=1

\documentclass[11pt]{article}

\usepackage{acl}

\usepackage{times}
\usepackage{latexsym}

\usepackage[T1]{fontenc}

\usepackage[utf8]{inputenc}

\usepackage{microtype}

\usepackage{inconsolata}

\usepackage{graphicx}
\usepackage{multirow}
\usepackage{url}
\usepackage{booktabs}
\usepackage{linguex}
\usepackage{amsmath} 
\usepackage{subfig}
\usepackage{tabularx}
\usepackage{enumitem}

\newcolumntype{Y}{>{\centering\arraybackslash}X}
\def\0{\phantom{0}} 
\def\dbl0{\phantom{0}\phantom{0}} 
\def\tpl0{\phantom{0}\phantom{0}\phantom{0}} 

\title{GDTB: Genre Diverse Data for English Shallow Discourse Parsing \\ across Modalities, Text Types, and Domains}

\author{Yang Janet Liu$^{2,3,\dagger}$\thanks{\;\;equal contribution; $^{\dagger}$work done while at Georgetown.}
\hspace{2pt}
Tatsuya Aoyama$^{1}$\footnotemark[1]
\hspace{2pt}
Wesley Scivetti$^{1}$\footnotemark[1]
\hspace{2pt}
Yilun Zhu$^{1}$\footnotemark[1]
\hspace{2pt} \\ 
{\bf Shabnam Behzad$^{1}$ Lauren Elizabeth Levine$^{1}$ Jessica Lin$^{1}$ Devika Tiwari$^{1}$ Amir Zeldes$^{1}$} \\
        $^1$Corpling Lab, Georgetown University \\ 
        $^2$MaiNLP, Center for Information and Language Processing, LMU Munich, Germany \\ $^3$Munich Center for Machine Learning (MCML)  \\
        {\tt y.liu1@lmu.de \hspace{2pt}
        \{ta571,wss37,yz565\}@georgetown.edu} \\ 
        {\tt \{sb1796,lel76,yl1290,dt719,az364\}@georgetown.edu}        
}

\begin{document}
\maketitle
\begin{abstract}
Work on shallow discourse parsing in English has focused on the Wall Street Journal corpus, the only large-scale dataset for the language in the PDTB framework. However, the data is not openly available, is restricted to the news domain, and is by now $35$ years old. In this paper, we present and evaluate a new open-access, multi-genre benchmark for PDTB-style shallow discourse parsing, based on the existing UD English GUM corpus, for which discourse relation annotations in other frameworks already exist. In a series of experiments on cross-domain relation classification, we show that while our dataset is compatible with PDTB, substantial out-of-domain degradation is observed, which can be alleviated by joint training on both datasets.
\end{abstract}

\section{Introduction}\label{sec:intro}

Language in discourse is more than an ordered list of sentences or clauses: parts of a text expressing events, states, facts, or propositions are often connected by discourse relations, such as \textsc{cause} (one part of a text specifies the cause for another), \textsc{concession} (one part states content which a speaker or author expects recipients to overlook) etc. Such relations may be marked \textit{explicitly}, for example by \textit{connectives}, which are conjunctions or adverbials such as `because' or `nevertheless' in English; or they may be \textit{implicit}, requiring recipients to interpret the text more actively.

Given an arbitrary natural language text as input, shallow discourse parsing is the task of identifying pairs of text spans connected by a discourse relation, in some scenarios focusing mainly on explicit, implicit, or other subtypes of relations
\cite{xue-etal-2016-conll}, as well as the means by which they are signaled, resulting for example in connective disambiguation (e.g.~`since' is a connective which can express either a \textsc{cause} or \textsc{temporal} relation). Most commonly, shallow discourse parsing systems use the inventory of relations defined by the Penn Discourse Treebank (PDTB, currently version 3; see Section \ref{subsec:frameworks}).

Systems and data for shallow discourse parsing can be used for a variety of downstream applications, including relation extraction (identification of relations given two spans, \citealt{braud-etal-2024-disrpt-multilingual}), instruction fine-tuning or pretraining of language models \cite{EinDorEtAl2022}, study of argumentation and persuasiveness \cite{rehbein-2019-role}, and cross-linguistic lexicography of discourse connectives \cite{scheffler-stede-2016-adding,das-etal-2018-constructing,kurfali-etal-2020-ted}. When finding specific relation types is desired, shallow discourse parsing also forms an end task in itself: for example, finding all \textsc{concession} relations in a large corpus of speeches by a politician or political party for Computational Social Science studies.

Although work on shallow discourse parsing has expanded to a range of languages (e.g.~Chinese, \citealt{Zhou2014TheCD}; Czech, \citealt{synkova-etal-2024-announcing-prague}, German, \citealt{sluyter-gathje-etal-2020-shallow}; Italian, \citealt{tonelli-etal-2010-annotation}; Thai, \citealt{PrasertsomEtAl2024}, Turkish, \citealt{zeyrek-kurfali-2017-tdb}, Nigerian Pidgin, \citealt{saeed2024implicitdiscourserelationclassification}), less progress has been made on expanding data to new and diverse domains (see Section \ref{subsec:datasets}). A major cause of this bottleneck is the effort associated with manual construction of high quality data covering a broad range of domains from scratch.

In this paper we suggest overcoming this hurdle by not starting from scratch:~we target the freely available English GUM corpus (which is also available as part of the Universal Dependencies project, \citealt{de-marneffe-etal-2021-universal}), which covers a broad range of $16$ spoken and written English genres and for which annotations are available in hierarchical discourse parsing frameworks:~RST and eRST (see Section \ref{subsec:frameworks}). Although these frameworks are substantially different from PDTB, they provide sufficient information to obtain a high quality starting point for semi-automatic conversion of data into the PDTB v3 framework. As an added advantage, we also develop a mapping of (e)RST to PDTB relations, allowing for cross-framework comparisons along the lines proposed by \citet{Demberg2019HowCA} (see \citealt{zhu-etal-2021-ontogum} for a similar argument and approach to converting coreference datasets).

In the subsequent sections of this paper, we will first briefly survey the discourse relation frameworks involved in this project (Section \ref{sec:related}), and then we will present our data, its creation process, and an evaluation of its quality (Section \ref{sec:GDTB}). This will be followed by a set of experiments on cross-corpus and joint-training relation classification to evaluate both the compatibility of our data with PDTB, and the degree of cross-corpus (and by extension, cross-domain) performance degradation.

\section{Related Work}\label{sec:related}

\subsection{Discourse Relation Frameworks}\label{subsec:frameworks}

A number of frameworks have been proposed for the computational modeling of discourse relations. The Penn Discourse TreeBank (PDTB; \citealp{prasad-etal-2014-reflections}), as briefly outlined above, is a lexically grounded shallow discourse parsing framework, which proposes that texts contain any finite amount of discourse relations (including possibly zero) from an inventory of $36$ relations (as of v3) presented in Appendix \ref{appendix:pdtb-v3-sense-inventory}, which hold between potentially overlapping spans of text.

PDTB's lexical grounding means that each relation is associated with a kind of triggering device allowing for its identification:~\textit{\textbf{explicit}} relations correspond to a (possibly multi-word) lexical item which in English is either a subordinating conjunction (`because'), a coordinating one (`but'), or an adverbial, including adverbs (`however') and prepositional phrases (`at the same time'). By contrast, \textit{\textbf{implicit}} relations are identified by the potential insertability of a connective, which is not actually present in the text, and generally hold either between consecutive sentences in the same paragraph, or between a small set of additional constructions (e.g.~purpose infinitives, for which we can insert an implicit `in order (to)'; see Section \ref{subsec:conversion}). PDTB further includes some relations using non-connective expressions: 

\begin{itemize}
    \item \textbf{alternative lexicalizations} (\textsc{AltLex}): non-connective words such as `this causes' (instead of `because')
    \item \textbf{alternative lexicalization constructions} (\textsc{AltLexC}): constructions with connective-like functions such as auxiliary inversion in `had I gone' (instead of `if')
    \item \textbf{entity relations} (\textsc{EntRel}): elaborating relations mediated by coreferring entities. 
    \item \textbf{hypophora}: the relation between questions and their answers 
\end{itemize}

\noindent Adjacent sentence pairs in the same paragraph not mediated by these relations are tagged as \textsc{NoRel}. Relations in PDTB are hierarchical (e.g.~\textsc{Comparison.Contrast} is a type of \textsc{Comparison}) and either symmetrical (e.g.~\textsc{Comparison.Similarity}) or specify a direction using a third level of hierarchy (e.g.~\textsc{Comparison.Concession.arg2-as-denier} specifies which argument span is being conceded).

The two other most prominent discourse relation frameworks for which substantial implemented corpora exist are Rhetorical Structure Theory (RST \citealt{MannThompson1988}) and Segmented Discourse Representation Theory (SDRT, \citealt{AsherLascarides2003}), 
which both assume that texts can be completely segmented into elementary discourse units (EDUs, roughly equivalent to propositions), and that EDUs always connect to form a hierarchical graph (in the case of RST, a projective, single rooted tree). Since English SDRT corpora are limited in genre and domain, primarily covering videogame chat \cite{asher-etal-2016-discourse,thompson-etal-2024-discourse-structure} and help forum discussions \cite{li-etal-2020-molweni}, we focus here on RST, which has been applied to a broad range of domains (see \citealt{da-cunha-etal-2011-development,liu-zeldes-2023-cant}) and languages (e.g.~Basque, \citealt{da-cunha-iruskieta-2010}, Chinese, \citealt{peng-etal-2022-gcdt}, Russian, \citealt{PisarevskayaEtAl2017} and more). An example of RST discourse annotation is illustrated for a text fragment in Figure \ref{fig:rst-erst-example} (disregarding blue edges and highlighted words, see below).

RST enforces a single tree hierarchical structure over an entire document, assuming that every smallest unit of analysis (i.e.~EDU) is related to another unit or subtree by one of the proposed discourse relations such as \textsc{cause}, \textsc{background}, or \textsc{contrast} (see Appendix \ref{appendix:rst-relation-inventory} for the relation inventory used in GUM). Importantly, such relations are annotated in the spirit of \textit{plausibility judgments} (\citealp[246]{MannThompson1988}) from the perspective of the writer, independent of the words in the text, meaning that it is fundamentally pragmatic in orientation, rather than lexically grounded. Relations in RST are either directed, from a less prominent satellite to a more prominent nucleus, or symmetrical, forming multinuclear units. However, RST does not mark connectives or other expressions indicative of relations, and is incapable of expressing multiple, concurrent relations on the same nodes, or tree-breaking relations.

\begin{figure}
    \centering
    \includegraphics[width=\columnwidth]{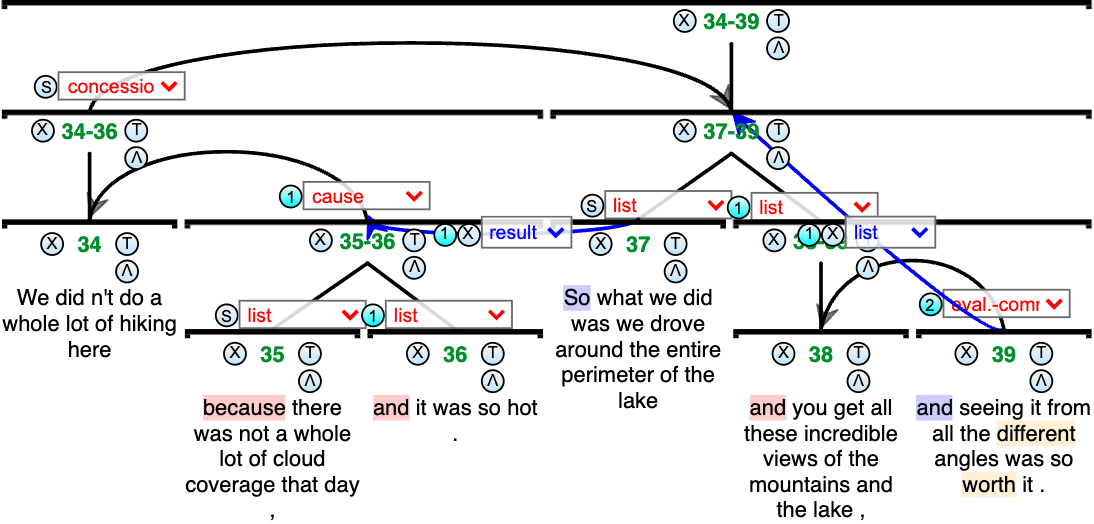}
    \caption{A Discourse Analysis in RST (disregarding blue edges and highlighted words) and eRST.}
    \label{fig:rst-erst-example}
\end{figure}

Recently, \citet{zeldesEtAl2024erst} proposed an enhanced version of RST called eRST, which adds additional, tree-breaking relations on top of RST, as well as signaling annotations, which indicate the rationale for relation annotations using $45$ signal categories arranged into $8$ classes. \citet{zeldesEtAl2024erst} also released a re-annotated version of the multi-genre English GUM RST treebank (see Section \ref{sec:GDTB}) containing enhanced RST trees with relation signals. Fortunately for the present purpose, one of the signal classes in this dataset corresponds to connectives based on PDTB definitions, facilitating the conversion process we will outline below. Figure \ref{fig:rst-erst-example} highglights the additional structures of the eRST graph, in which tree-breaking relations are marked by \textcolor{blue}{blue} edges, and signal token spans are highlighted in different colors based on their classes. Connectives corresponding to regular relations are marked in \textcolor{red}{red}, and those corresponding to tree-breaking edges are marked in \textcolor{blue}{blue}.

\subsection{Datasets}\label{subsec:datasets}

Due to its size, work on shallow discourse parsing in English has focused on PDTB, which, as of version 3, contains over $53$K discourse relation instances, the majority of which are either explicitly or implicitly associated with connectives (about $24$K and $21$K each). Despite its impressive size, PDTB is limited as a resource for text covering language other than newswire, since its underlying data comes exclusively from the Wall Street Journal (WSJ) corpus, containing WSJ articles from 1989. As a result, the data is missing both a range of contemporary thematic content (e.g.~words like `cellphone', `European Union', or `website' are absent) and diverse modes of communication (e.g.~PDTB's \textsc{Hypophora} relation, indicating question-answer pairs, occurs only $141$ times, and spoken language connectives such as `cause'/`cuz' are unattested).

The few other available datasets which cover English discourse relations in the PDTB-style are newer, but much smaller, and cover few domains: The TED Multilingual Discourse Bank (TED-MDB, \citealt{zeyrek2019ted}) follows PDTB in including not only explicit and implicit relations, but also \textsc{AltLex}, \textsc{EntRel}, and \textsc{NoRel} annotations for TED talks translated into six languages, for a total of $3,649$ relations. However, only $661$ of those annotations cover English data, and the corpus has not been updated to conform to the PDTB v3 guidelines.

A much larger dataset, Edina-DR \cite{ma-etal-2019-implicit}, contains $27,998$ implicit relations from conversational data. However, the corpus is annotated fully automatically and only at the top level of the PDTB relation hierarchy, thereby distinguishing only $4$ relation labels (\textsc{Comparison}, \textsc{Expansion}, \textsc{Contingency}, and \textsc{Temporal}).

DiscoGeM \cite{scholman-etal-2022-discogem} contains $6,505$ relations from Wikipedia texts, European Parliament proceedings and literature, and comes closest to the goal of the resource presented here in offering diverse text types with detailed manual relation annotations. Version 2.0 of the corpus \cite{yung-etal-2024-discogem-2} also adds parallel data for a subset of relations for three languages: Czech, French, and German. However, the corpus only covers inter-sentential \textit{implicit} relations, thereby limiting the scope of the task substantially, and still contains no conversational spoken data, academic writing, YouTube data etc., which we aim to cover with GDTB. With this in mind we present the contents of our corpus in the next section.

\section{GDTB}\label{sec:GDTB}

\subsection{Contents}

\begin{table}[h!t]
\centering
\small
\resizebox{5.5cm}{!}{%
\begin{tabular}{l|rr}
    \toprule
     & \textbf{GDTB} & \textbf{PDTB v3}\\
    \midrule
    Tokens & 228,399 & 1,156,308 \\
    Docs & \tpl0 235 & \tpl0 2,161 \\
    Genres & 16 & 1 \\
    \midrule
    AltLex & 224 & 1,498 \\
    AltLexC & 13 & 140 \\
    EntRel & 553 & 5,538 \\
    Explicit & 7,202 & 24,238 \\
    Hypophora & 465 & 146 \\
    Implicit & 4,503 & 21,781 \\
    Norel & 662 & 287 \\
    \midrule
    All & 13,622 & 53,628 \\
        \bottomrule
\end{tabular}
}
\caption{Relation Type Counts: GDTB vs.~PDTB v3.}
\vspace{-5pt}
\label{tab:corp-overview}
\end{table}

The dataset presented in this paper is the GUM Discourse Treebank (GDTB), a multi-genre PDTB v3-style corpus for English semi-automatically converted from the GUM corpus \cite{Zeldes2017}.\footnote{Our data is made available at \url{https://github.com/gucorpling/gum2pdtb} under a Creative Commons license in accordance with the original GUM license. Data from the Reddit genre \cite{behzad-zeldes-2020-cross} is released without underlying text, but a script is provided to reconstruct the data using an API. We plan to include future versions of GDTB with new documents as part of the main GUM corpus releases, as the GUM corpus grows.} GUM is a growing multilayer corpus of English containing, among other things, discourse parses with aligned connective annotations in eRST, Universal Dependencies syntax trees, entity and coreference annotations, and more.

After conversion of the data, the process for which is described below, the final GDTB benchmark based on GUM v10 contains $13.6$K relation annotations, a little more than a quarter of the size of PDTB v3, but stemming from much more diverse and up to date materials. Table \ref{tab:corp-overview} compares the two datasets. 
Note that because sentences in some genres are shorter than in news text, GDTB is less than $1/4$ the size of PDTB in tokens, but denser in discourse relations; at the same time, shorter paragraphs in many genres mean the proportion of implicit relations is lower. Some relation types are also more frequent in GDTB; in particular, \textsc{Hypophora}, which corresponds to questions, is common in many genres but rare in newspaper language. The underlying data in GUM is regularly expanded and currently covers $16$ genres, where data collection for four of these is still ongoing (`growing' genres). Table \ref{tab:genre-overview} gives an overview of the data, with the four growing genres at the bottom.
\begin{table}[h!tb]
\centering
\small
\resizebox{7cm}{!}{%
\begin{tabular}{l|rrr}
\toprule
Genre & Docs & Tokens & Relations \\
\midrule
\textit{academic} & 18 & 17,169 & 815 \\
\textit{bio} & 20 & 18,213 & 868 \\
\textit{conversation} & 14 & 16,391 & 1,113 \\
\textit{fiction} & 19 & 17,510 & 1,281 \\
\textit{interview} & 19 & 18,196 & 1,188 \\
\textit{news} & 23 & 16,146 & 724 \\
\textit{reddit} & 18 & 16,364 & 1,146 \\
\textit{speech} & 15 & 16,720 & 913 \\
\textit{textbook} & 15 & 16,693 & 936 \\
\textit{vlog} & 15 & 16,864 & 1,415 \\
\textit{voyage} & 18 & 16,514 & 799 \\
\textit{how-to} & 19 & 17,081 & 1,331 \\
\midrule
\textit{court} & 6 & 7,069 & 478 \\
\textit{essay} & 5 & 5,750 & 348 \\
\textit{letter} & 6 & 5,982 & 365 \\
\textit{podcast} & 5 & 5,737 & 359 \\
\bottomrule
\end{tabular}
}
\caption{Genre Breakdown for GDTB. The bottom four `growing' genres are still being collected for GUM and counts represent sizes as of GUM v10.}
\vspace{-8pt}
\label{tab:genre-overview}
\end{table}
Genres cover both spoken (e.g.~conversations, courtroom transcripts, YouTube vlogs) and written modalities (incl.~news, academic, how-to guides from wikiHow) from various open licensed sources, which should make models trained on GDTB more robust to open domain data (see Section \ref{sec:experiments}).

\subsection{Dataset Conversion}\label{subsec:conversion}

As mentioned above, GUM v10 contains several annotation layers that describe linguistic phenomena at various levels, including eRST trees, but also gold syntax and coreference annotations, which we harness to create GDTB. 

\paragraph{Sense Mapping} Our approach to creating GDTB uses a cascade of relation conversion modules, and manual annotation for some types of error-prone cases in the entire corpus (all relations in the test set are also manually annotated). All modules rely on a mapping of allowable output relations, adapted from the PDTB v2 proposal in \citet{Demberg2019HowCA}, which had to be modified in several ways (see also \citealt{costa-etal-2023-mapping} on mapping v3 data). PDTB v3 introduced finer-grained Level-3 sense distinctions, which are mostly concerned with relation directionality. Because what PDTB calls \textit{Arg1} and \textbf{Arg2} is determined by the syntactic configuration and their linear order in the text, its interaction with the order-dependent Level-3 senses is not straightforwardly mappable from RST relations, where directionality is based on labels (e.g.~\textsc{cause} vs.~\textsc{result}) and nuclearity or relative prominence. That said, in many cases a deterministic mapping can be achieved (e.g.~what an RST \textsc{concession} relation concedes is reliably the opposite argument span of the \textsc{.argX-as-denier} argument in PDTB v3).

Secondly, the RST framework adopted in \citet{Demberg2019HowCA} is based on the RST-DT corpus \cite{carlson2003building}, which uses a set of relation labels slightly different from that of GUM v10. This incompatibility was resolved in consultation with the original RST-DT relation descriptions \cite{carlson2003building} and the description of GUM v10 discourse relations.\footnote{\url{https://wiki.gucorpling.org/gum/rst}} Finally, since the resulting label mapping is still often many-to-many, each module employs different strategies to  disambiguate potential PDTB senses, which are detailed below. Figure \ref{fig:gdtb-conversion-examples} in Appendix \ref{appendix:gdtb-conversion-examples} presents some GDTB examples spawned by RST annotations given our conversion process described below.

\paragraph{Explicit Module} Explicit relation candidates are generated with simple heuristics: (1) for each eRST relation in GUM, add the relation to the candidate list if it is signaled by a connective; (2) for each relation in the candidate list, determine the allowable PDTB labels based on the connective and the RST relation; and (3) take the target and source EDU spans and convert them into PDTB argument spans based on another set of rules (see \textbf{Argument Span Module} below for more details).

For step (1), we use the gold GUM eRST framework annotations from \citet{zeldesEtAl2024erst} to determine if a given relation is explicitly signaled by a connective. For (2), we refer to Appendix \ref{appendix:pdtb-v3-sense-inventory} of the PDTB v3 annotation guidelines \citep{webber2019penn} to obtain a list of corresponding connectives and PDTB senses they may signal. For simplicity, rare combinations, such as secondary senses (e.g.~\textsc{Temporal.Synchronous|Comparison.Contrast}) and speech act variants \textsc{XYZ+SpeechAct} (which only make up $0.4$\% of the PDTB explicit data, or $121$ cases) are not considered. For cases where multiple outcomes are possible, we train DisCoDisCo \cite{gessler-etal-2021-discodisco}, a discourse relation classification system which remains state-of-the-art on the DISRPT shared task benchmark for relation classification \cite{braud-etal-2024-disrpt-multilingual}, on PDTB v3, and use its predictions to disambiguate data in our training and development sets. The test set is completely manually corrected to allow for the evaluation in Section \ref{sec:experiments}.

\paragraph{Implicit Module} Implicit relations are also handled in a three-step approach: (1) identify every junction allowing an implicit relation (viz.~between sentences, before purpose infinitives and participial adverbial clauses, and between zero-coordinated clauses); (2) predict the connective given existing RST relations, and (3) map the connective and relation onto a PDTB relation. For (1) we use the gold syntax trees and RST relations, which allow us, for example, to identify sentence boundaries, purpose infinitives, etc. Implicit relations are only allowed in the absence of an explicit relation in the same span, with one exception: RST \textsc{sequence} relations signaled by an explicit \textsc{Expansion.Conjunction} connective (e.g.~`and'), are allowed a second \textsc{Temporal} relation with implicit `then', matching PDTB's policy, as in example \ref{ex:and-then}. 

\ex. I cut my losses \textbf{and} (\textit{then}) ran - `and' \\ {\small Expl. \textsc{Expansion.Conjunction}} + `(then)'  \\ {\small Impl. \textsc{Temporal.Asynchronous.Precedence}} \label{ex:and-then}

For step (2), we train a connective prediction model to output a list of hypothetical connectives for each relation. This model is trained on implicit relations from the PDTB training set. We also supply the model with information about the possible PDTB relation senses that are compatible with existing RST relations at that juncture as part of the input. Specifically, we fine-tune \texttt{flan-t5-large}\footnote{\url{https://huggingface.co/google/flan-t5-large}} \citep{chung2024scaling} for $25$ epochs for this task and select the best-performing model on the dev set. See Appendix \ref{appendix:implicit-conn-prediction} for task performance and comparison to a majority baseline, and Appendix \ref{appendix:implicit-prompt-examples} for example prompts used in this task. 
%

We manually validate the entire test set and establish that the process is generally reliable for well-mappable relations (e.g.~RST \textsc{condition} or \textsc{cause} relations are easy to map), but less reliable for RST relations with no specific equivalents. In particular, we identify a high error rate for RST \textsc{context-background} and \textsc{joint-other}, which we manually correct for the entire corpus. 

Following connective prediction we use the same mapping of connectives and RST relation combinations as the explicit module to select the most likely PDTB relation. In ambiguous cases we again rely on DisCoDisCo predictions, except for relations corresponding to RST \textsc{context-background} and \textsc{joint-other} relations, or in the test set, which we manually correct.

\paragraph{AltLex Module} In some instances, there is no explicit connective present to signal a discourse relation, but the insertion of an implicit connective appears redundant due to an existing expression in the spans under consideration. In such cases, PDTB recognizes this expression as an \textit{alternative lexicalization} (\textsc{AltLex}) of the discourse relation, annotating the span of the \textsc{AltLex} expression, its relation, and its corresponding argument spans.

As there is no specific syntactic requirements for \textsc{AltLex} expressions, their detection is challenging. For this corpus conversion, we adopt a conservative approach for the annotation of potential \textsc{AltLex} expressions, adopting a pattern-matching approach similar to that outlined in \citet{knaebel-stede-2022-towards}, catching only cases which are attested in PDTB v3. 
The \textsc{AltLex} module is only consulted in the absence of an Explicit relation.

\paragraph{AltLexC Module} \textsc{AltLexC} is a subtype of \textsc{AltLex}, where the relation is expressed by syntactic constructions within a sentence, as shown in \ref{ex-aux-inv}. The \textsc{AltLexC} module is only consulted in the absence of an Explicit or \textsc{AltLex} relation. The list of acknowledged constructions for \textsc{AltLexC} in PDTB v3 is closed, making a rule-based approach based on syntax trees straightforward. To identify these, we first extract the seven syntactic constructions listed in the PDTB v3 annotation guidelines \citep{webber2019penn}, such as Auxiliary Inversion, where \textbf{Arg2} signals the \textsc{Contingency.Condition} sense. 

\ex. \label{ex-aux-inv}
\textbf{\underline {Had it happened} five hours earlier or four hours earlier}, \textit{I think the death toll would have been more than a thousand.}

We then verify that the matching syntax trees are contained in spans connected with compatible RST relations based on the mapping. For example, the original RST relation \textsc{contingency-condition} in \ref{ex-aux-inv} is mapped onto the PDTB relation \textsc{Contingency.Condition}. Although this process only captures $13$ instances of \textsc{AltLexC} in the corpus, these were error free based on manual inspection, and additional searches in the syntax annotations suggest that these have been exhaustively identified in the corpus.

\paragraph{Hypophora Module} The hypophora relation type is straightforwardly generated for each RST \textsc{topic-question} relation in the source annotations, since these correspond exactly to questions, the category covered by hypophora.

\paragraph{EntRel Module} At the juncture of each two adjacent, same-paragraph sentences for which no other relation has been generated, we check the GUM gold coreference and RST annotations to see whether any kind of \textsc{Joint} or \textsc{Elaboration} relation applies, and if so, whether it corresponds to coreference from a definite or pronominal expression in the second sentence referring back to the first sentence. If so, we generate an \textsc{EntRel} relation, and otherwise, mark the span as \textsc{NoRel}, following PDTB guidelines. This annotation type is manually corrected only in the test set.

\paragraph{Argument Span Module} In order to make target and source EDU spans conform to PDTB-style argument spans, we first apply the argument labeling convention described in the PDTB v3 annotation manual (\citealp[Section 3.1]{webber2019penn}) to each pair of RST EDU spans, which are a set of rules based on syntactic configuration and linear text order. Once the argument labeling has determined which span is \textit{Arg1}/\textbf{Arg2}, we refine the sense labels by adding the Level-3 sense information to restore directionality. 
In addition, we emulate PDTB's \textit{Minimality Principle}, which states that only the minimal text needed for a given discourse relation should be included in the argument spans \cite{prasad-etal-2014-reflections}. As a result, we also adjust the corresponding EDU spans by clipping EDU spans to a single sentence that contains just the head EDU if they are multi-sentential, and the exact span dominated by a relation source or target intra-sententially. Attribution spans which scope over an argument nucleus are also removed, in accordance with PDTB guidelines (e.g.~\textit{[X said] [A happened] [because B]} results in the removal of `X said' for the \textsc{cause} relation). Argument span accuracy is evaluated below.

\subsection{Evaluation}\label{subsec:eval}

\begin{table}[htb]
\centering
\small
\resizebox{7cm}{!}{%
\begin{tabular}{l|ccc}
\toprule
\multicolumn{4}{l}{\textbf{Relation Scores (exact label and span match)}} \\
\midrule
\textbf{type} & \textbf{P} & \textbf{R} & \textbf{F1} \\
\midrule
\textbf{altLex} & 0.9500 & 0.7600 & 0.8444 \\
\textbf{altLexC} & 1.0000 & 1.0000 & 1.0000 \\
\textbf{EntRel} & 0.7593 & 0.8913 & 0.8200 \\
\textbf{Explicit} & 0.9812 & 0.9874 & 0.9843 \\
\textbf{Hypophora} & 0.8750 & 0.8537 & 0.8642 \\
\textbf{Implicit} & 0.8784 & 0.8205 & 0.8485\\
\textbf{NoRel} & 0.7887 & 0.9180 & 0.8485 \\
\midrule
\textit{\textbf{micro-avg.}} & 0.9277 & 0.9161 & \textbf{0.9218} \\
\midrule
\multicolumn{4}{l}{\textbf{Span Scores (incl.~relation type but not sense)}} \\
\midrule
\textbf{altLex} & 0.9500 & 0.7600 & 0.8444 \\
\textbf{altLexC} & 1.0000 & 1.0000 & 1.0000 \\
\textbf{EntRel} & 0.7778 & 0.9130 & 0.8400 \\
\textbf{Explicit} & 0.9935 & 1.0000 & 0.9967 \\
\textbf{Hypophora} & 0.8750 & 0.8537 & 0.8642 \\
\textbf{Implicit} & 0.9824 & 0.9176 & 0.9489 \\
\textbf{NoRel} & 0.7887 & 0.9180 & 0.8485 \\
\midrule
\textit{\textbf{micro-avg.}} & 0.9678 & 0.9554 & \textbf{0.9616} \\
\bottomrule
\end{tabular}
}
\caption{Test Set Accuracy (manual correction).}
\vspace{-5pt}
\label{tab:manual-acc}
\end{table}

To assess the quality of the annotations in GDTB, we evaluate system outputs against the manually corrected test set ($1531$ relations), as well as conducting an inter-annotator agreement study. For the first experiment, we compare the fully corrected test data to the same test data but with only corrections done on the entire corpus, such as inspection of \textsc{background} and \textsc{other} relations. Following an initial training session with joint adjudication, data was corrected by a team of nine Computational Linguistics graduate students and faculty with formal training in discourse parsing formalisms as part of a research project. Following previous work we evaluate on Level-2 relations in two scenarios: \textbf{\textit{exact match}}, where the label, argument span, and relation type must match, and \textbf{\textit{span-only match}}, meaning the relation type was identified and argument spans are correct but the label may not be.

As Table \ref{tab:manual-acc} shows, the overall quality of the corpus is very high, with a micro-F1 score of $92$, above our initial expectations given human agreement scores reported for PDTB annotation. Although there are no comprehensive numbers available for PDTB v3 annotation, \citet{prasad-etal-2008-penn} reported $84$\% accuracy (exact match between annotators) on v2 senses, which did not include the more challenging intra-sentential implicit relations, \textsc{AltLexC}, or Hypophora, and \citet{zeyrek2019ted} similarly reported $79$\% agreement. 
\citet{bourgonje-stede-2020-potsdam} reported a Cohen’s Kappa of $0.74$ on all relations, again excluding intra-sentential relations.\footnote{We considered reporting kappa for our data as well, but this requires the set of relations for annotation to match and carry only one label each. Using argument spans to align instances, we can compute kappa for the $93.7$\% of relations which have exactly one sense in both the predicted and corrected data -- for these we obtain $\kappa$=$0.913$.}
On the lower end, \citet{scholman-etal-2022-discogem} reported $60$\% agreement and $\kappa$=.$45$ on Level-3 senses using the v2 inventory with crowd workers, while \citet{yung-etal-2024-discogem-2} emphasized the importance of collecting multiple labels and reporting confusion matrices.

As the bottom half of Table \ref{tab:manual-acc} shows, argument spans are relatively unproblematic compared to sense prediction, especially for implicit cases, where span matching achieves almost $0.95$, but exact match F1 including sense is just below $0.85$. This is not unexpected, given that human judgments on insertion of an unexpressed connective vary considerably. For more details on the kinds of labels that human annotators corrected, and detailed confusion matrices, see Appendix \ref{appendix:confmat-corrections}.
%
%
We further double-annotated $8$ documents from the test set focusing on implicit instances. A Cohen's kappa of $0.79$ was achieved on connectives, $\kappa$=$0.77$ on Level-3 senses, and $\kappa$=$0.83$ on Level-2 using PDTB v3 inventory, indicating excellent agreement.

\section{Experimental Setup}\label{sec:experiments}

To test the utility of our corpus and its compatibility (or redundancy) with the existing PDTB v3, we again train the DisCoDisCo relation classifier \cite{gessler-etal-2021-discodisco} using the standard DISRPT version of PDTB v3 relations, which simply provides the spans of the two arguments including connectives, and their containing sentences, without a separate connective field, and as a result treats explicit, implicit, and other relation types uniformly,\footnote{DISRPT datasets do not contain instances of \textsc{EntRel}.} though we also report separate scores on different relation types and overall. 

To investigate the effects of both data size and data diversity, we evaluate in three training setups: \textbf{within-corpus} (e.g.~train and test on PDTB v3, and the same for GDTB respectively); \textbf{cross-corpus} (train on PDTB v3 and evaluate on the GDTB test set, and vice versa); and \textbf{joint training} (train on both training sets, evaluate on each test set). See implementation details in Appendix \ref{appendix:implementation}.

\section{Results}\label{sec:results}

Table \ref{tab:nlp-result-overall} gives an overview of within-, across-, and joint-corpus overall relation classification accuracy. The overall scores show that GDTB is the more challenging corpus when training is done jointly, and cross-corpus degradation is non-negligible, with around $10$ points degradation for training on PDTB v3 and testing on GDTB, and even more so in the opposite direction. 
\begin{table}[ht]
\centering
\resizebox{5cm}{!}{%
\begin{tabular}{@{}l|cc@{}}
\toprule
  & \multicolumn{2}{c}{\textbf{Test Set}} \\
\midrule
           \textbf{Training}             & \textbf{GDTB} & \textbf{PDTB v3} \\ \midrule
\textbf{within-corpus}  & 0.6447        & 0.7572           \\
\textbf{cross-corpus}   & 0.5660        & 0.4457           \\
\textbf{joint-training} & 0.6440        & 0.7390           \\ \bottomrule
\end{tabular}%
}
\caption{Overall Accuracy Scores (within-corpus=train set is from the corpus of the test set; cross-corpus=train set from opposite corpus; joint=train on both).}
\label{tab:nlp-result-overall}
\end{table}
\begin{table}[h!t]
\centering
\resizebox{\columnwidth}{!}{%
\begin{tabular}{@{}l|l|ccccc@{}}
\toprule
\textbf{Train} & \textbf{Test} & \textbf{Explicit} & \textbf{Implicit} & \textbf{altLex} & \textbf{altLexC} & \textbf{Hypophora} \\ \midrule
\multirow{2}{*}{\textbf{GDTB}} & GDTB & 0.7645 & 0.4579 & 0.4400 & 1 & 0.8780 \\
 & PDTB v3 & 0.6114 & 0.2842 & 0.3333 & 0.5000 & 0.7500 \\ \midrule
\multirow{2}{*}{\textbf{PDTB v3}} & GDTB & 0.6794 & 0.4048 & 0.3600 & 1 & 0.5854 \\
 & PDTB v3 & 0.8817 & 0.6020 & 0.8986 & 0.9167 & 0.8750 \\ \midrule
\multirow{2}{*}{\textbf{\begin{tabular}[c]{@{}l@{}}GDTB \&\\ PDTB v3\end{tabular}}} & GDTB & 0.7374 & 0.4908 & 0.4400 & 1 & 0.9512 \\
 & PDTB v3 & 0.8679 & 0.5683 & 0.8261 & 0.8333 & 0.8750 \\ \bottomrule
\end{tabular}%
}
\caption{Accuracy by Relation Types.}
\label{tab:nlp-result-everything}
\end{table}
\noindent Although joint training slightly under-performs within-corpus numbers in both directions, the relatively low level of degradation for the joint model compared to the corresponding within-corpus numbers suggests that the joint model is a much better choice for training a system to tag truly unseen, open domain data. 

However, the different proportions of explicit relations (which are easier to tag) and implicit ones mean that Table \ref{tab:nlp-result-overall} does not give the entire picture. Thus, we also report accuracy scores for different relation types in Table \ref{tab:nlp-result-everything}, which shows that the relation type which benefits most from joint training is \textsc{Hypophora}, which is rare in PDTB v3; without GDTB data, the PDTB-trained model degrades $29$ points out of domain. 
For \textsc{Hypophora}, we see that the joint model out-performs or performs as well as single corpus training and testing. The joint model is unsurprisingly superior in the cross-corpus macro-average, which is a better proxy for realistic applications to unseen data in the wild.

Implicit relations in particular show massive cross-corpus degradation for PDTB, which is again unsurprising: while connectives remain more or less constant across datasets (i.e.~`but' usually signals \textsc{Comparison.Contrast} or \textsc{Comparison.Concession} in both datasets), in implicit settings, the relations between surrounding lexical items must be learned, which vary more substantially by genre. Although the GDTB model has seen some news data from GUM \textit{news}, the quantity of the material is insufficient to achieve comparable scores to the PDTB-trained model, which has $1.2$M tokens of WSJ data to learn from.

\begin{figure}[h!t]
    \centering
    \subfloat[within-corpus model.]{\includegraphics[width=\columnwidth, scale=1]{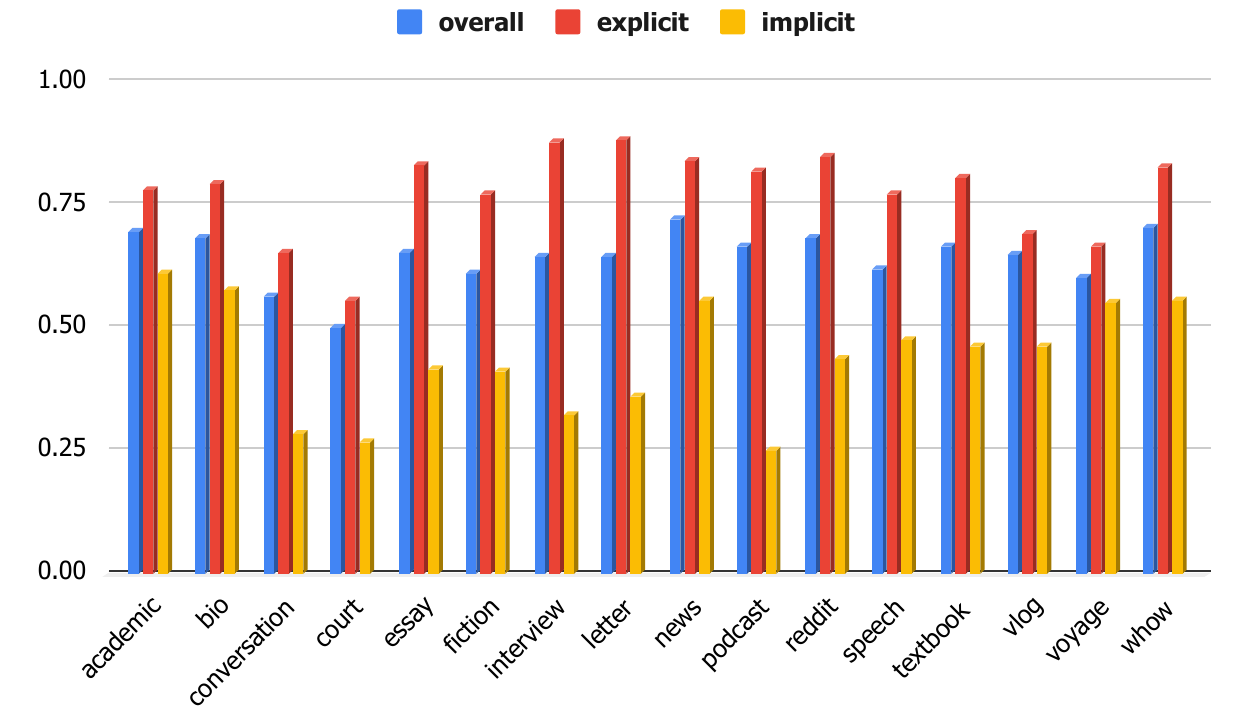}\vspace{-8pt}}\label{fig:within-corpus-gdtb-test}
    \centering
    \subfloat[cross-corpus model.]{\includegraphics[width=\columnwidth, scale=1]{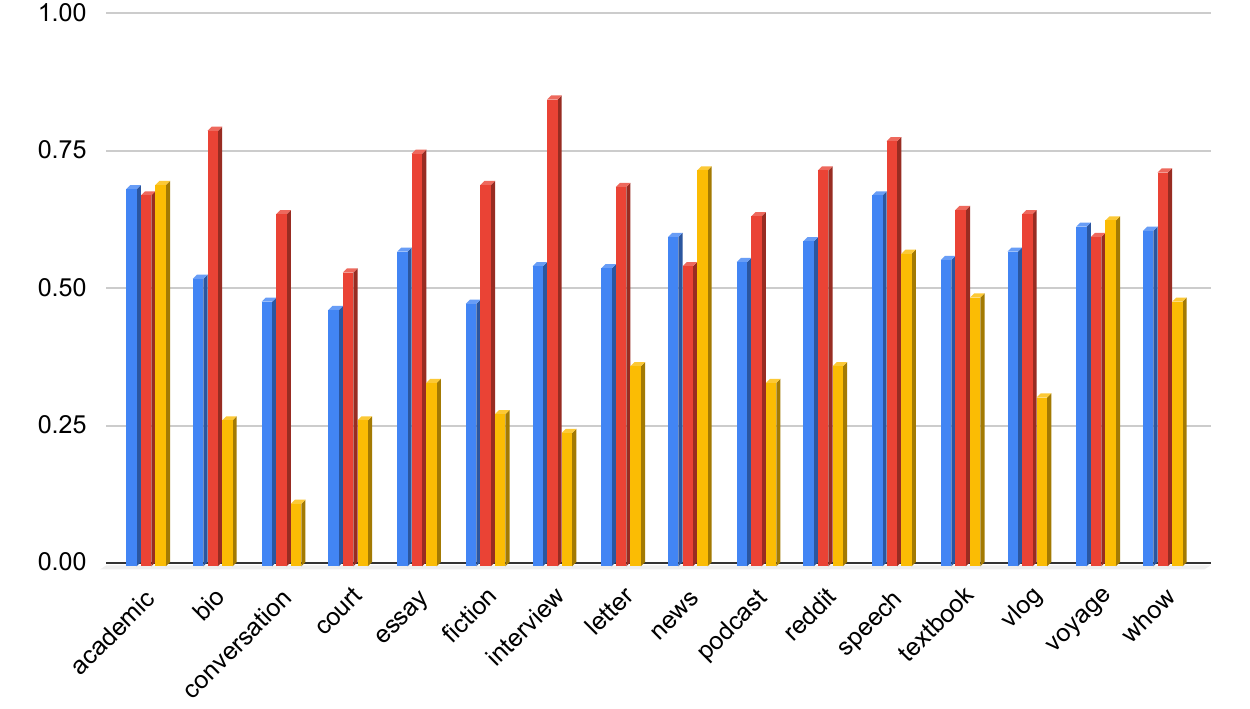}}\label{fig:cross-corpus-gdtb-test}
    \caption{GDTB Scores by Genres and Relation Types.}
    \label{fig:gdtb-test-by-genre}
\end{figure}

If we zoom in on genres in GDTB, Figure \ref{fig:gdtb-test-by-genre} shows scores for subsets of GDTB \texttt{test} from both the within-corpus and the cross-corpus models.\footnote{Full results and confusion matrices are in Appendix \ref{appendix:more-scores}.} Overall, the best-performing genres for each model are \textit{news} and \textit{academic} respectively, while the worst-performing genre is \textit{court} for both models. The bottom four places in implicit relations within-corpus are all occupied by spoken genres (\textit{podcast, court, conversation, interview}). The \textit{conversation} genre is particularly bad for cross-corpus, i.e.~when training on PDTB, with an implicit score of just $11.43$\%. Unsurprisingly, \textit{news} scores highest on cross-corpus implicit relation prediction ($72.22$\%), though \textit{academic} scores higher overall ($68.60$\%) (due to frequent explicit relations), 
and \textit{interview} scores highest for cross-corpus explicit ($84.85$\%), likely due to the frequent use of easy connectives such as \textsc{temporal} `when' in questions, and `then' in sequential narration in answers.

One of the major motivations of this work is releasing genre-diverse, complete PDTB v3-style data to facilitate cross-framework and cross-domain shallow discourse parsing. While, strictly speaking, there is no other dataset annotated like PDTB v3 to evaluate on, other existing datasets do contain a subset of PDTB v3 annotations. 
\begin{table}[h!t]
\resizebox{\columnwidth}{!}{%
\begin{tabular}{@{}l|ccc@{}}
\toprule
 & \textbf{GDTB-trained} & \textbf{PDTB-trained} & \textbf{joint-training} \\ \midrule
\multicolumn{1}{c|}{\textbf{\begin{tabular}[c]{@{}c@{}}TED-MDB\\ (English)\end{tabular}}} & 0.5214 & 0.5556 & 0.5641 \\ \bottomrule
\end{tabular}%
}
\caption{Accuracy Scores of TED-MDB (English).}
\label{tab:result-tedmdb-eng}
\end{table}
\noindent For instance, Table \ref{tab:result-tedmdb-eng} shows the accuracy scores of the English portion of the TED-Multilingual Discourse Bank (TED-MDB, \citealt{zeyrek-etal-2018-multilingual,ted-mdb}) across three training scenarios. As the table shows, we observe out-of-domain gains on the TED talks in the corpus when using the jointly trained model. However, it is worth pointing out that since TED-MDB only contains a subset of PDTB v3 annotations, these scores are not directly comparable to the scores reported in the tables above.

\section{Conclusion and Outlook}\label{sec:outlook}

In this paper, we present GDTB, a PDTB-style dataset covering $16$ English spoken and written genres for open-domain shallow discourse parsing, which we create primarily using a cascade of conversion modules leveraging enhanced RST annotations. The data covers all aspects of PDTB v3 annotation, including explicit and implicit inter-/intra-sentential relations as well as alternative lexicalizations, entity relations, and hypophora.

We show that RST relations lead to reliable PDTB-style annotations, particularly for explicit relations. Using state-of-the-art fine-tuned sequence-to-sequence models and the (e)RST relations as inputs, we are also able to obtain high quality predictions for implicit relations, which we correct in whole for the test set and in part for the remaining data's most unreliably convertible RST relation types (e.g.~\textsc{context-background} and \textsc{joint-other}).

Our experiments show that there is substantial degradation in cross-corpus PDTB-style relation classification in both directions, demonstrating PDTB's current inadequacy for relation classification in open domain settings. We show that jointly training a relation classification system on both PDTB v3 and GDTB leads to much greater cross-corpus stability without sacrificing much performance on PDTB v3. We are therefore confident that GDTB can be a valuable resource for improving out-of-domain performance of PDTB-style English shallow discourse parsing systems.

In future work, we believe the same pipeline presented in this paper can be applied to additional corpora annotated with RST in general and eRST in particular. Specifically, the recent addition of eRST annotations to the GENTLE corpus (\textbf{GEn}re \textbf{T}ests for \textbf{L}inguistic \textbf{E}valuation), an extension corpus applying GUM's annotation scheme to $8$ more unusual English genres \cite{aoyama-etal-2023-gentle}, should allow for more GDTB-like data to be produced with ease. This data would cover the additional GENTLE genres, which encompass dictionary entries, eSports video commentary, legal documents, medical notes, poetry, mathematical proofs, course syllabuses, and threat letters.

Finally, beyond discourse parsing, we also see great promise in using GDTB for theoretical studies of discourse relation variation across genres, and for the comparison of alignments between PDTB-style relations and RST or eRST annotations. In particular, we believe GDTB and GUM-RST will prove to be informative for future comparisons between theoretical frameworks, along the lines proposed by \citet{Demberg2019HowCA}.

\section*{Limitations}
\label{sec:limitations}

In this work, we use the gold eRST relations present in GUM as a starting point for our PDTB-style annotations. It is likely that annotating the same underlying data from scratch in the PDTB style would yield a slightly different result than our approach here, particularly regarding implicit relations. In PDTB, implicit relations are posited between all adjacent sentences but only between adjacent sentences which have an eRST relation that connects the two sentences in GDTB. This means that while implicit relation precision in GDTB has high quality, our recall is likely to be partial. We believe that due to the prioritization of pragmatically prominent relations in RST and the possibility of multiple tree-breaking relations in eRST, the most salient relations in documents should already be included in our data. In support of this claim, we note that previous work has found that PDTB-style relations for Czech seldom violate RST tree projectivity constraints \cite{PolakovaMirovskyZikanovaEtAl2021}, especially if multiple concurrent relations are permitted (see also \citealt{polakova-etal-2024-developing-rhetorical} on annotating RST relations for Czech PDTB-style data).

Another limitation is the noise inherent to a conversion approach which uses automatic processes. In our case, this is especially true for the automatically generated implicit connectives, which may diverge somewhat from the most natural connectives chosen by humans. As stated previously, this issue is particularly problematic for implicit relations spawned from RST relations without good PDTB mappings, such as \textsc{joint-other} and \textsc{context-background}, which were therefore manually corrected in our entire dataset (including correction of connectives), next to the correction of all connectives in the test set. That said, our evaluation shows that, by relying on multiple sources of information for our final predictions, the final product is substantially better than an automatically created dataset tagging just discourse relations from plain text, producing a resource that is close to gold-standard quality, or at a minimum, significantly `better-than-silver' (cf.~\citealt{gessler-etal-2020-amalgum}). Future work could improve the quality of additional predicted connectives to bring the complete corpus closer to gold-standard accuracy.

In addition, this work explores the possibility of converting an English RST-style discourse treebank to a PDTB-style one based on the English PDTB v3, leading to several limitations in the applicability of our methods. Primarily, the work is limited to English as a target language, and does not address the lack of diverse data in other languages. That being said, we believe that the resource created here is valuable for facilitating multilingual shallow discourse parsing, as recently experimented in \citet{bourgonje-demberg-2024-generalizing} where state-of-the-art model originally developed for English discourse relation classification was extended to a multilingual setting, and by employing some simple yet effective learning techniques, the discourse relation classification performance becomes more generalizable and robust across both languages and domains.

Lastly, our methods assume the existence of RST and connective annotations for the source material. While there are many RST datasets which could be converted following the methods proposed here, nearly no RST dataset also contains connective annotations (the German PCC corpus, \citealt{bourgonje-stede-2020-potsdam}, is an exception). Nevertheless, our methods could be applied to other RST corpora using automatic or manual connective annotation, especially in languages for which connective lexical resources and/or some PDTB-style data exist.

\section*{Ethics Statement}
\label{sec:ethics-statements}

This work, like most work in Computational Linguistics, can enable the creation of Natural Language Processing systems which may cause harm. However, we believe that the lack of diverse data for training systems is a greater potential harm than making additional data available, which will hopefully allow systems to behave in less biased and more generalizable ways.

In addition, this work employs deep learning architectures whose training involves carbon emissions. While these should not be ignored, we assess them to be of a modest scope, given that we are relying on existing pre-trained models, which are only fine-tuned on small amounts of data using limited resources. Finally, all human labor involved in this paper was carried out by paid university employees, including funded graduate students as part of their research work. No unpaid volunteers or low-paid crowd workers were involved in the creation of this data.

\section*{Acknowledgments}

We recognize the support for Yang Janet Liu through the ERC Consolidator Grant DIALECT 101043235.

\bibliography{anthology,custom}

\appendix

\section{PDTB v3 Sense Hierarchy}
\label{appendix:pdtb-v3-sense-inventory}

\begin{table}[ht]
\centering
\resizebox{\columnwidth}{!}{%
\begin{tabular}{@{}l|l|l@{}}
\toprule
\textbf{Level-1} & \textbf{Level-2} & \textbf{Level-3} \\ \midrule\midrule
\multirow{3}{*}{\textsc{temporal}} & \textsc{synchronous} & -- \\ \cmidrule(l){2-3} 
 & \multirow{2}{*}{\textsc{asynchronous}} & \textsc{precedence} \\ 
 &  & \textsc{succession} \\ \midrule\midrule
\multirow{15}{*}{\textsc{contingency}} & \multirow{3}{*}{\textsc{cause}} & \textsc{reason} \\
 &  & \textsc{result} \\
 &  & \textsc{negResult} \\ \cmidrule(l){2-3} 
 & \multirow{2}{*}{\textsc{cause+belief}} & \textsc{reason+belief} \\
 &  & \textsc{result+belief} \\
 & \multirow{2}{*}{\textsc{cause+SpeechAct}} & \textsc{reason+SpeechAct} \\
 &  & \textsc{result+SpeechAct} \\ \cmidrule(l){2-3} 
 & \multirow{2}{*}{\textsc{condition}} & \textsc{arg1-as-cond} \\
 &  & \textsc{arg2-as-cond} \\ \cmidrule(l){2-3} 
 & \textsc{condition+SpeechAct} & -- \\ \cmidrule(l){2-3} 
 & \multirow{2}{*}{\textsc{negative-condition}} & \textsc{arg1-as-negCond} \\
 &  & \textsc{arg2-as-negCond} \\ \cmidrule(l){2-3} 
 & \textsc{negative-condition+SpeechAct} & -- \\ \cmidrule(l){2-3} 
 & \multirow{2}{*}{\textsc{purpose}} & \textsc{arg1-as-goal} \\
 &  & \textsc{arg2-as-goal} \\ \midrule\midrule
\multirow{5}{*}{\textsc{comparison}} & \multirow{2}{*}{\textsc{concession}} & \textsc{arg1-as-denier} \\
 &  & \textsc{arg2-as-denier} \\ \cmidrule(l){2-3} 
 & \textsc{concession+SpeechAct} & \textsc{arg2-as-denier+SpeechAct} \\ \cmidrule(l){2-3} 
 & \textsc{contrast} & -- \\ \cmidrule(l){2-3} 
 & \textsc{similarity} & -- \\ \midrule\midrule
\multirow{13}{*}{\textsc{expansion}} & \textsc{conjunction} & -- \\ \cmidrule(l){2-3} 
 & \textsc{disjunction} & -- \\ \cmidrule(l){2-3} 
 & \textsc{equivalence} & -- \\ \cmidrule(l){2-3} 
 & \multirow{2}{*}{\textsc{exception}} & \textsc{arg1-as-excpt} \\
 &  & \textsc{arg2-as-excpt} \\ \cmidrule(l){2-3} 
 & \multirow{2}{*}{\textsc{instantiation}} & \textsc{arg1-as-instance} \\
 &  & \textsc{arg2-as-instance} \\ \cmidrule(l){2-3} 
 & \multirow{2}{*}{\textsc{level-of-detail}} & \textsc{arg1-as-detail} \\
 &  & \textsc{arg2-as-detail} \\ \cmidrule(l){2-3} 
 & \multirow{2}{*}{\textsc{manner}} & \textsc{arg1-as-manner} \\
 &  & \textsc{arg2-as-manner} \\ \cmidrule(l){2-3} 
 & \multirow{2}{*}{\textsc{substitution}} & \textsc{arg1-as-subst} \\
 &  & \textsc{arg2-as-subst} \\ \bottomrule 
\end{tabular}%
}
\caption{An Overview of the PDTB v3 Sense Hierarchy.}
\vspace{-10pt}
\label{tab:pdtb-v3-inventory}
\end{table}

Table \ref{tab:pdtb-v3-inventory} presents the PDTB v3 Sense Hierarchy, reproduced from the PDTB v3 annotation manual (\citealp[Section 4]{webber2019penn}). This work reproduces the complete Level-3 sense labels, with the exception of the rare \textsc{+Belief} and \textsc{+SpeechAct} variants, which are collapsed to their corresponding basic Level-3 variants.

\section{RST Relation Inventory in GUM}
\label{appendix:rst-relation-inventory}

Table \ref{tab:gum-rst-inventory} presents the RST relation inventory used in GUM, both fine-grained relation labels as well as the corresponding coarse-grained relation classes are provided. Note that \textsc{same-unit} is not a true discourse relation but instead a label used to mark discontinuous spans in RST as a result of RST's EDU segmentation. We include it here for completeness purposes. 

\begin{table}[ht]
\centering
\resizebox{\columnwidth}{!}{%
\begin{tabular}{@{}l|l|l|l@{}}
\toprule
\textbf{\begin{tabular}[c]{@{}l@{}}GUM v10 \\ Classes\end{tabular}} & \textbf{\begin{tabular}[c]{@{}l@{}}GUM v10\\ Relations\end{tabular}} & \textbf{\begin{tabular}[c]{@{}l@{}}GUM v10 \\ Classes\end{tabular}} & \textbf{\begin{tabular}[c]{@{}l@{}}GUM v10\\ Relations\end{tabular}} \\ \midrule
\multirow{3}{*}{\textsc{Adversative}} & \textsc{adversative-antithesis} & \multirow{4}{*}{\textsc{Joint}} & \textsc{joint-disjunction} \\
 & \textsc{adversative-concession} &  & \textsc{joint-list} \\
 & \textsc{adversative-contrast} &  & \textsc{joint-sequence} \\ \cmidrule(r){1-2}
\multirow{2}{*}{\textsc{Attribution}} & \textsc{attribution-positive} &  & \textsc{joint-other} \\ \cmidrule(l){3-4} 
 & \textsc{attribution-negative} & \multirow{2}{*}{\textsc{Mode}} & \textsc{mode-manner} \\ \cmidrule(r){1-2}
\multirow{2}{*}{\textsc{Causal}} & \textsc{causal-cause} &  & \textsc{mode-means} \\ \cmidrule(l){3-4} 
 & \textsc{causal-result} & \multirow{3}{*}{\textsc{Organization}} & \textsc{organization-heading} \\ \cmidrule(r){1-2}
\multirow{2}{*}{\textsc{Context}} & \textsc{context-background} &  & \textsc{organization-phatic} \\
 & \textsc{context-circumstance} &  & \textsc{organization-preparation} \\ \midrule
\textsc{Contingency} & \textsc{contingency-condition} & \multirow{2}{*}{\textsc{Purpose}} & \textsc{purpose-attribute} \\ \cmidrule(r){1-2}
\multirow{2}{*}{\textsc{Elaboration}} & \textsc{elaboration-attribute} &  & \textsc{purpose-goal} \\ \cmidrule(l){3-4} 
 & \textsc{elaboration-additional} & \multirow{2}{*}{\textsc{Restatement}} & \textsc{restatement-partial} \\ \cmidrule(r){1-2}
\multirow{3}{*}{\textsc{Explanation}} & \textsc{explanation-evidence} &  & \textsc{restatement-repetition} \\ \cmidrule(l){3-4} 
 & \textsc{explanation-justify} & \multirow{2}{*}{\textsc{Topic}} & \textsc{topic-question} \\
 & \textsc{explanation-motivation} &  & \textsc{topic-solutionhood} \\ \midrule
\textsc{Evaluation} & \textsc{evaluation-comment} & \textsc{same-unit} & \textsc{same-unit} \\ \bottomrule
\end{tabular}%
}
\caption{RST Relation Inventory in GUM v10.}
\vspace{-8pt}
\label{tab:gum-rst-inventory}
\end{table}

\section{Implicit Connective Prediction Performance}
\label{appendix:implicit-conn-prediction}

Regarding the evaluation of implicit connective predictions, we calculated two types of accuracy of predicted connective in GDTB against the human evaluations on the GDTB test set: \textbf{\textit{exact match}} and \textbf{\textit{fuzzy match}}. Overall, for exact match the connectives were judged as natural by our evaluators at an accuracy of $79$\%. While accuracy at matching the connective exactly is somewhat low, many of the model predicted connectives were compatible with the correct PDTB sense, but were changed by annotators to add additional fluency and naturalness. The system predicts a reasonable connective, that is, a connective that is valid for the gold PDTB v3 sense, at an accuracy of $89$\%. We call this more lenient scoring method the \textit{fuzzy match} accuracy, compared to the \textit{exact match} accuracy where the connective must be identical to what the annotator ultimately decided on. We also report a genre breakdown for both scoring scenarios in Table \ref{tab:impl-conn-pred-acc-exact-and-fuzzy}. Overall, the higher \textit{fuzzy match} scores indicate that the vast majority of automatically generated connectives are at least reasonable, even though annotators sometimes decide that a more natural sounding connective is possible.

For comparison, we also compute a majority baseline for each RST relation in GDTB test. The baseline predicts the most frequent connective given the RST relation that spawns the GDTB relation. We find that this majority baseline has an exact match score of $51$\%. We find that the majority baseline produces a fuzzy match score of $88$\%, indicating that the RST relation is a very strong signal towards the set of PDTB-valid connectives. However, we find that our model provides substantially more natural connectives, leading to higher exact match scores. A genre breakdown for the majority baseline is reported in Table \ref{tab:impl-conn-pred-acc-exact-and-fuzzy-majority} below.

\begin{table}[ht]
\centering
\small
\resizebox{6cm}{!}{%
\begin{tabular}{@{}l|cc@{}}
\toprule
\textbf{Genres} & \textbf{\begin{tabular}[c]{@{}c@{}}Exact Match\\ Accuracy\end{tabular}} & \textbf{\begin{tabular}[c]{@{}c@{}}Fuzzy Match\\ Accuracy\end{tabular}} \\ \midrule
\textit{academic} & 0.710 & 0.871 \\
\textit{bio} & 0.814 & 0.907 \\
\textit{conversation} & 0.885 & 0.962 \\
\textit{court} & 0.846 & 0.923 \\
\textit{essay} & 0.727 & 0.818 \\
\textit{fiction} & 0.679 & 0.839 \\
\textit{interview} & 0.735 & 0.882 \\
\textit{letter} & 0.900 & 0.950 \\
\textit{news} & 0.882 & 0.941 \\
\textit{podcast} & 0.875 & 0.875 \\
\textit{reddit} & 0.811 & 0.865 \\
\textit{speech} & 0.634 & 0.829 \\
\textit{textbook} & 0.756 & 0.854 \\
\textit{vlog} & 0.958 & 0.958 \\
\textit{voyage} & 0.789 & 0.921 \\
\textit{how-to} & 0.878 & 0.939 \\ \bottomrule
\end{tabular}%
}
\caption{Exact and Fuzzy Match Accuracy of Connective Prediction by Genres (Fine-tuned Connective Prediction Model).}
\label{tab:impl-conn-pred-acc-exact-and-fuzzy}
\end{table}

\begin{table}[h!t]
\centering
\small
\resizebox{6cm}{!}{%
\begin{tabular}{@{}l|cc@{}}
\toprule
\textbf{Genres} & \textbf{\begin{tabular}[c]{@{}c@{}}Exact Match\\ Accuracy\end{tabular}} & \textbf{\begin{tabular}[c]{@{}c@{}}Fuzzy Match\\ Accuracy\end{tabular}} \\ \midrule
\textit{academic} & 0.548 & 0.967 \\
\textit{bio} & 0.558 & 0.930\\
\textit{conversation} & 0.731 & 1.0 \\
\textit{court} & 0.308 & 0.846 \\
\textit{essay} & 0.318 & 0.818 \\
\textit{fiction} & 0.500 & 0.928 \\
\textit{interview} & 0.441 & 0.853 \\
\textit{letter} & 0.550 & 0.800 \\
\textit{news} & 0.471 & 0.882 \\
\textit{podcast} & 0.250 & 0.750 \\
\textit{reddit} & 0.567 & 0.865 \\
\textit{speech} & 0.537 & 0.756 \\
\textit{textbook} & 0.463 & 0.854 \\
\textit{vlog} & 0.625 & 0.958 \\
\textit{voyage} & 0.447 & 0.895\\
\textit{how-to} & 0.612 & 0.878 \\ \bottomrule
\end{tabular}%
}
\caption{Exact and Fuzzy Match Accuracy of Connective Prediction by Genres (Majority Baseline).}
\vspace{-10pt}
\label{tab:impl-conn-pred-acc-exact-and-fuzzy-majority}
\end{table}

\section{Implicit Connective Prediction Prompt}
\label{appendix:implicit-prompt-examples}

Table \ref{tab:implicit-prompt-examples} provides example prompts that were used to train Flan-T5 for the connective prediction task. The selected examples come from implicit relations in the training split of PDTB v3 used in the DISRPT shared task \cite{braud-etal-2023-disrpt}, which are what the model was trained on. For consistency, we label the argument spans as `Sentence 1' and `Sentence 2', regardless of whether the relation is inter- or intra-sentential.

\begin{table*}[ht]
\centering
\resizebox{13.5cm}{!}{%
\begin{tabular}{p{17cm}|p{2cm}}
\toprule
\textbf{Input} & \textbf{Output} \\ \midrule
Sentence 1: In July , the Environmental Protection Agency imposed a gradual ban on virtually all uses of asbestos . Sentence 2: By 1997 , almost all remaining uses of cancer-causing asbestos will be outlawed . Relations: contingency.cause.reason,contingency.purpose,contingency.cause.result &
  as a result \\ \midrule
Sentence 1: Sales figures of the test-prep materials are n't known , but their reach into schools is significant . Sentence 2: In Arizona , California , Florida , Louisiana , Maryland , New Jersey , South Carolina and Texas , educators say they are common classroom tools . Relations: contingency.condition,comparison.contrast,expansion.level-of-detail,expansion.manner,expansion.conjunction,expansion.instantiation,contingency.negative-condition &
  for example \\ \midrule
Sentence 1: Choose 203 business executives , including , perhaps , someone from your own staff , Sentence 2: and put them out on the streets , Relations: temporal.asynchronous.precedence,temporal.synchronous,temporal.asynchronous.succession,expansion.conjunction &
  then \\ \bottomrule
\end{tabular}%
}
\caption{Example Prompts used for the Connective Prediction Task.}
\vspace{-8pt}
\label{tab:implicit-prompt-examples}
\end{table*}

\section{Corrected Relation Correspondences}
\label{appendix:confmat-corrections}

\begin{figure}[ht]
    \centering
    \includegraphics[width=\columnwidth]{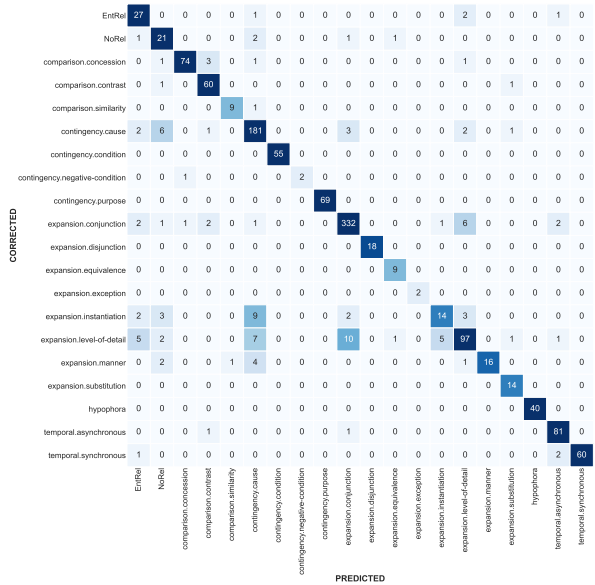}
    \caption{Confusion Matrix for Corrected Relations and their Initially Predicted Labels.}
    \vspace{-10pt}
    \label{fig:confmat-corrections}
\end{figure}

Figure \ref{fig:confmat-corrections} provides the confusion matrix for manually corrected relations in the test set (corrected relations on the y-axis) and their originally predicted relation as outputted by the conversion process (x-axis). The figure indicates that, apart from errors being overall rare, some relation predictions are completely reliable (e.g.~Hypophora are trivial to predict given RST's \textsc{question-answer} annotations). The most frequent error type is inferring a \textsc{cause} relation where annotators felt a less marked \textsc{Expansion} was warranted. Generally speaking, correction into an \textsc{Expansion} category was the most common type, likely because implicit \textsc{Expansion} connectives, such as `and', `in fact', or `specifically' are among the easiest to insert between sentence pairs.

\section{Implementation Details}
\label{appendix:implementation}

The connective prediction task within the implicit module was conducted using NVIDIA RTX A6000 GPUs with $64$GB RAM. Experiments related to DisCoDisCo's relation classification task were conducted on $1$ NVIDIA Tesla L4 GPU with $24$GB GPU Memory on Google Cloud Platform. For DiscoDisCo, overall we followed the original hyperparameters and training settings therein.\footnote{\url{https://github.com/gucorpling/DisCoDisCo}} However, we did not use any hand-crafted features proposed in the original work as such features are not available for GDTB and show degradation for PDTB v3, according to \citet{gessler-etal-2021-discodisco}.

\section{Examples of GDTB based on the Conversion Process}
\label{appendix:gdtb-conversion-examples}

\begin{figure*}[ht]
    \centering
    \includegraphics[width=\textwidth]{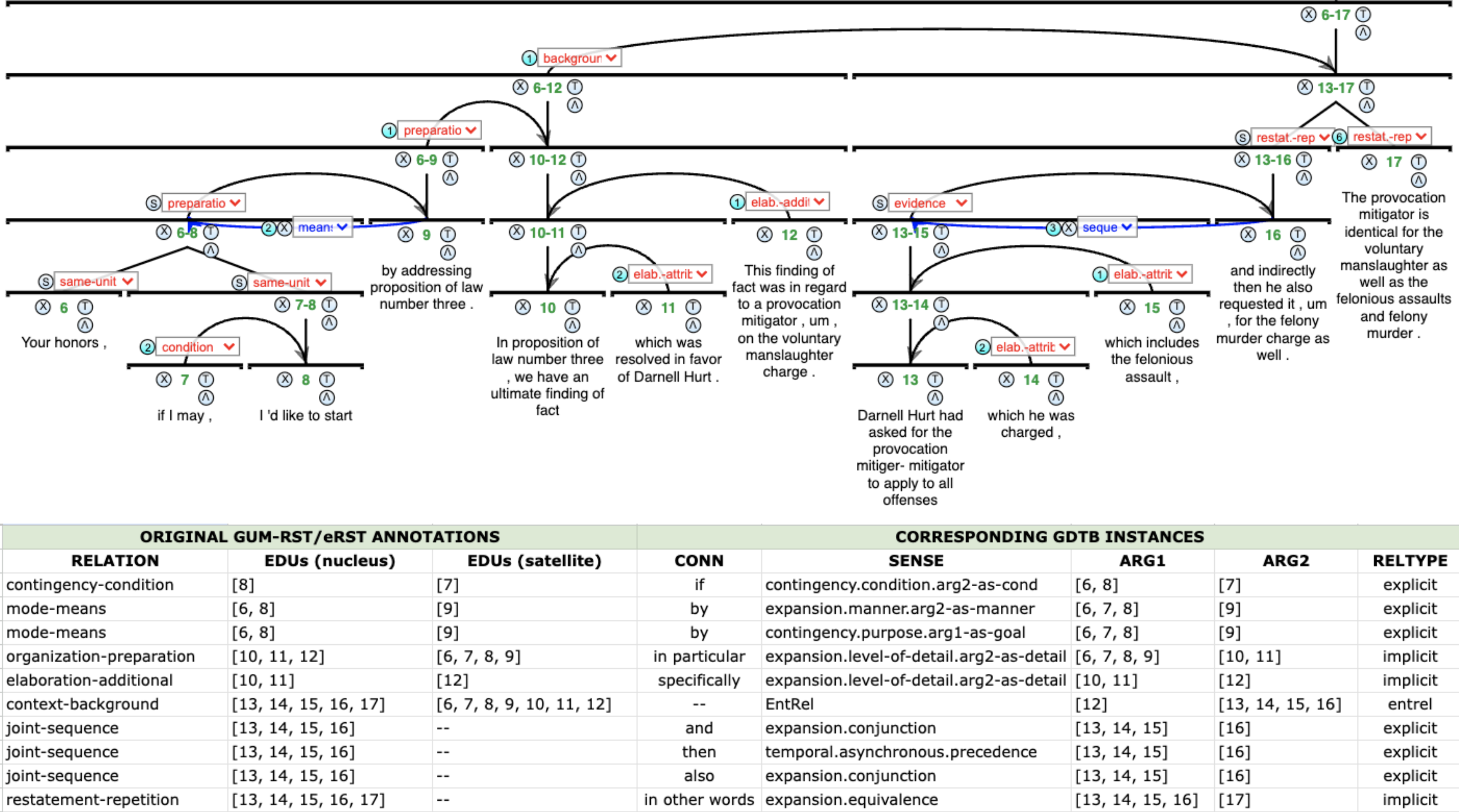}
    \caption{Examples of GDTB based on Gold GUM-RST Annotations and the Corresponding PDTB-style Instances.}
    \vspace{-8pt}
    \label{fig:gdtb-conversion-examples}
\end{figure*}

Figure \ref{fig:gdtb-conversion-examples} provides an illustration of an RST fragment from a document in GUM as well as a set of the PDTB-style relations spawned from the gold RST annotations given the conversion process described in Section \ref{subsec:conversion}, covering explicit, implicit, and entity relations. ``--'' in the satellite EDU column indicates that the relation is multi-nucleus, where both EDU spans are considered nuclei (such relations still have $2$ EDU spans, from which PDTB-style \textit{Arg1} and \textbf{Arg2} are spawned). ``--'' in the connective column means that the relation was not signaled by connectives (and was either signaled by other signals or unsignaled).

\section{Full Results and Confusion Matrices}
\label{appendix:more-scores}

\begin{figure*}[!ht]
    \centering
    \subfloat[Overall.]{\includegraphics[width=\columnwidth]{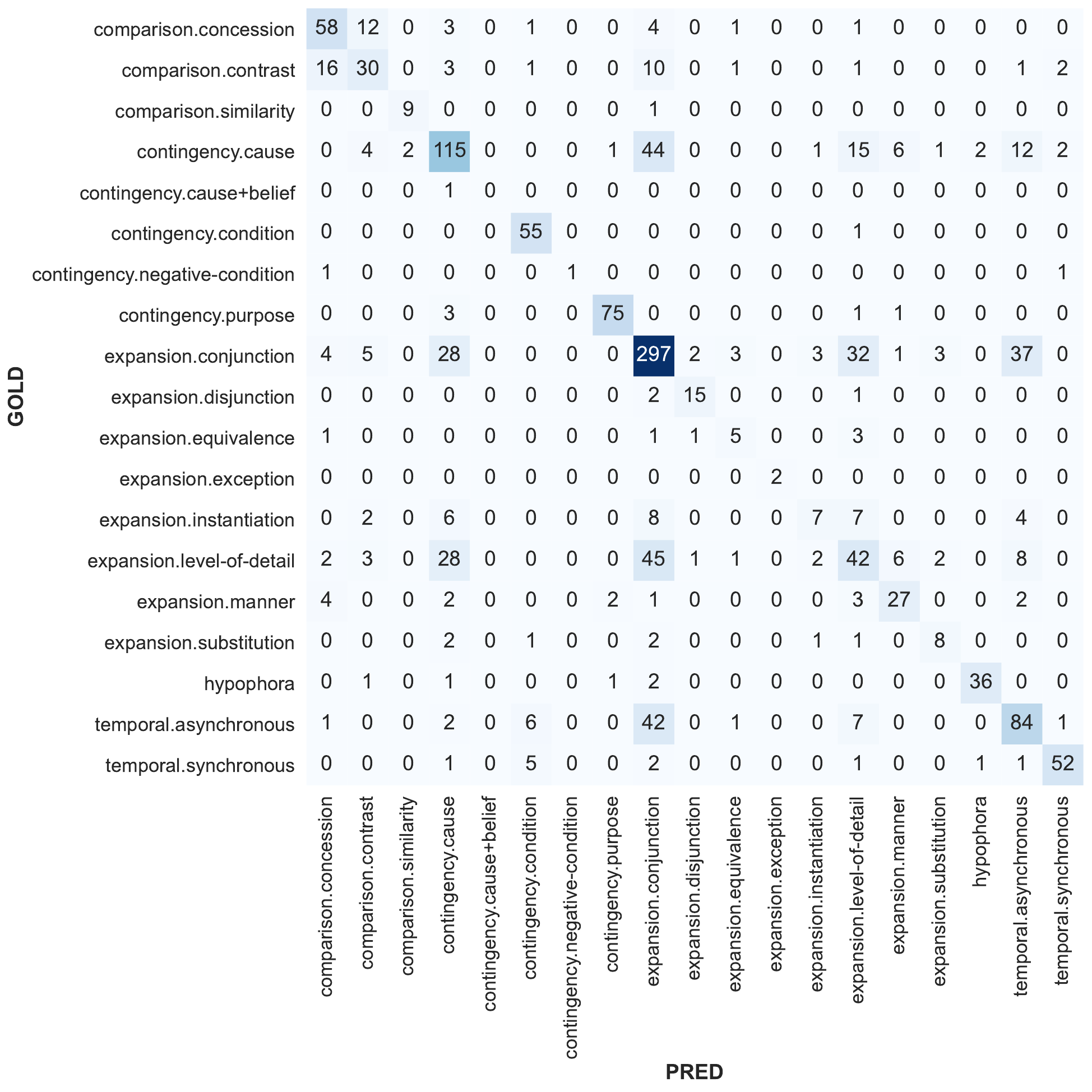}}
    \label{fig:gdtb-overall}
    \quad
    \centering
    \subfloat[Explicit Relations.]{\includegraphics[width=\columnwidth,scale=1]{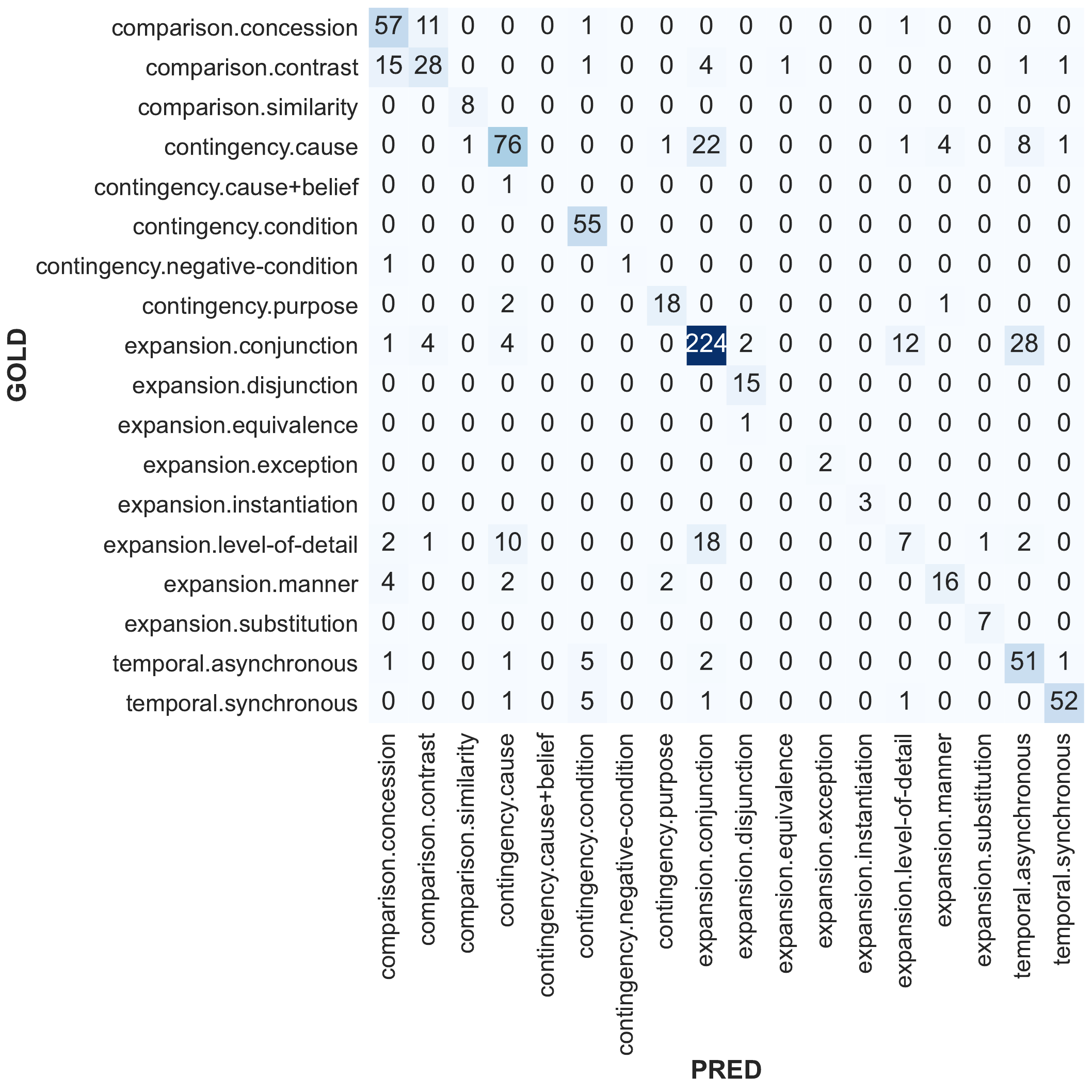}}\label{fig:gdtb-confusion-matrix-explicit}
    \quad 
    \centering
    \subfloat[Implicit Relations.]{\includegraphics[width=\columnwidth,scale=1]{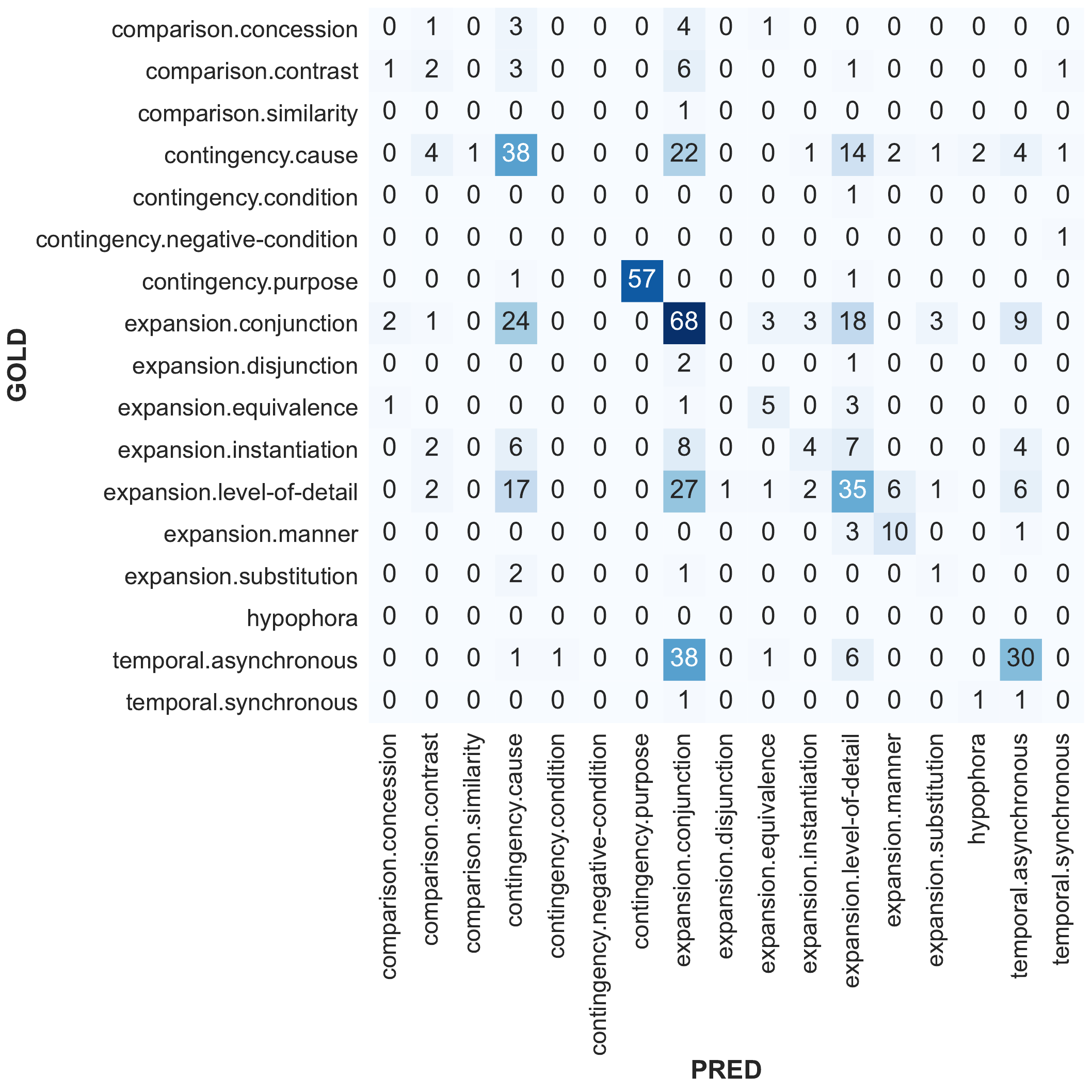}}\label{fig:gdtb-confusion-matrix-implicit}
    \quad 
    \centering
    \subfloat[AltLex Relations.]{\includegraphics[width=\columnwidth,scale=1]{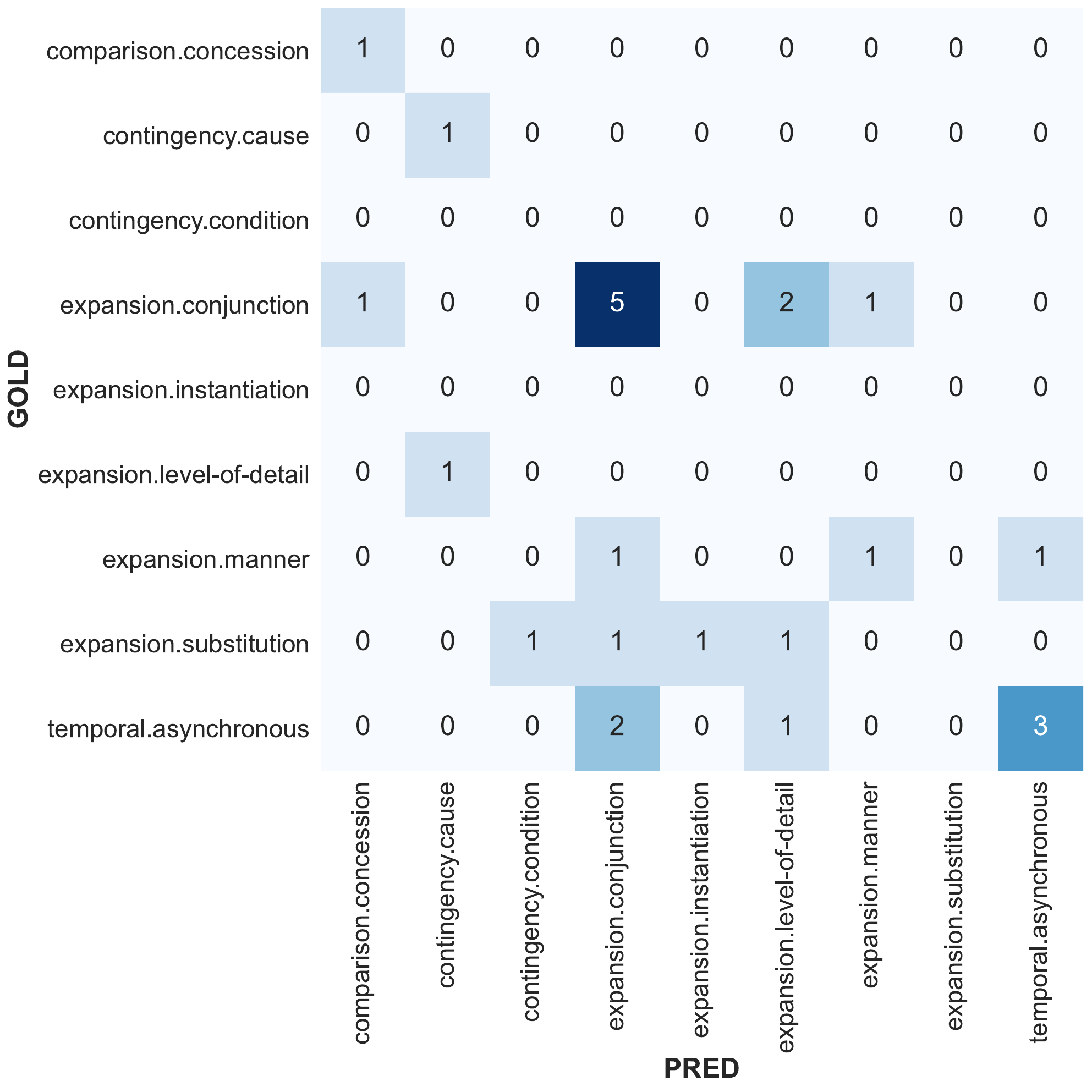}}\label{fig:gdtb-confusion-matrix-altlex}
    
    \caption{Confusion Matrices for GDTB \texttt{test} from the within-corpus Experiment.}
    \label{fig:gdtb-within-corpus-confmats}
\end{figure*}

Table \ref{tab:gdtb-scores} presents the accuracy of the GDTB test set from both the within-corpus and cross-corpus models by genres and relation types. Since some relation types such as \textsc{AltLex} and \textsc{Hypophora} are very rare, their scores are not available. 
In addition, we provide four confusion matrices in Figure \ref{fig:gdtb-within-corpus-confmats} that give a better idea of what PDTB v3 sense labels are prone to errors overall as well as for the major relation types including explicit, implicit, and \textsc{AltLex} relations. Unsurprisingly, \textsc{Expansion.Conjunction} is often the sense label that models tend to overpredict. In particular, it is commonly confused with \textsc{Contingency.Cause} across the board, but it tends to be more easily confused with \textsc{Temporal.Asynchronous} and \textsc{Expansion.Level-of-detail} for implicit relations. For explicit relations, \textsc{Expansion.Conjunction} is often confused with \textsc{Expansion.Level-of-detail}.

\begin{table*}[ht]
\resizebox{\textwidth}{!}{%
\begin{tabular}{@{}l|llllll|c|lllllll@{}}
\toprule
 & \multicolumn{6}{c|}{\textbf{within-corpus}} &  & \multicolumn{7}{c}{\textbf{cross-corpus}} \\ \cmidrule(r){1-7} \cmidrule(l){9-15} 
\multicolumn{1}{c|}{} & \multicolumn{1}{c}{\textbf{overall}} & \multicolumn{1}{c}{\textbf{explicit}} & \multicolumn{1}{c}{\textbf{implicit}} & \multicolumn{1}{c}{\textbf{altLex}} & \multicolumn{1}{c}{\textbf{altLexC}} & \multicolumn{1}{c|}{\textbf{hypophora}} &  & \multicolumn{1}{c|}{\textbf{}} & \multicolumn{1}{c}{\textbf{overall}} & \multicolumn{1}{c}{\textbf{explicit}} & \multicolumn{1}{c}{\textbf{implicit}} & \multicolumn{1}{c}{\textbf{altLex}} & \multicolumn{1}{c}{\textbf{altLexC}} & \multicolumn{1}{c}{\textbf{hypophora}} \\ \cmidrule(r){1-7} \cmidrule(l){9-15} 
\textit{academic} & 0.6977 & 0.7826 & 0.6111 & 0.5 & -- & -- &  & \multicolumn{1}{l|}{\textit{academic}} & 0.686 & 0.6739 & 0.6944 & 0.75 & -- & -- \\
\textit{bio} & 0.6818 & 0.7949 & 0.5778 & 0.6667 & 1 & -- &  & \multicolumn{1}{l|}{\textit{bio}} & 0.5227 & 0.7949 & 0.2667 & 0.6667 & 1 & -- \\
\textit{conversation} & 0.5669 & 0.6533 & 0.2857 & 0 & -- & 0.8667 &  & \multicolumn{1}{l|}{\textit{conversation}} & 0.4803 & 0.64 & 0.1143 & 0 & -- & 0.6 \\
\textit{court} & 0.5 & 0.5581 & 0.2667 & -- & -- & 1 &  & \multicolumn{1}{l|}{\textit{court}} & 0.4667 & 0.5349 & 0.2667 & -- & -- & 0.5 \\
\textit{essay} & 0.6557 & 0.8333 & 0.4167 & 0 & -- & -- &  & \multicolumn{1}{l|}{\textit{essay}} & 0.5738 & 0.75 & 0.3333 & 0 & -- & -- \\
\textit{fiction} & 0.6098 & 0.7742 & 0.4138 & 1 & -- & 1 &  & \multicolumn{1}{l|}{\textit{fiction}} & 0.4797 & 0.6935 & 0.2759 & 0 & -- & 0 \\
\textit{interview} & 0.6477 & 0.8788 & 0.3243 & 0.3333 & -- & 1 &  & \multicolumn{1}{l|}{\textit{interview}} & 0.5455 & 0.8485 & 0.2432 & 0.3333 & -- & 0.6667 \\
\textit{letter} & 0.6458 & 0.8846 & 0.3636 & -- & -- & -- &  & \multicolumn{1}{l|}{\textit{letter}} & 0.5417 & 0.6923 & 0.3636 & -- & -- & -- \\
\textit{news} & {\color[HTML]{333333} 0.72} & 0.8387 & 0.5556 & 0 & -- & -- &  & \multicolumn{1}{l|}{\textit{news}} & 0.6 & 0.5484 & 0.7222 & 0 & -- & -- \\
\textit{podcast} & 0.6667 & 0.8182 & 0.25 & -- & -- & -- &  & \multicolumn{1}{l|}{\textit{podcast}} & 0.5556 & 0.6364 & 0.3333 & -- & -- & -- \\
\textit{reddit} & 0.6833 & 0.8472 & 0.439 & 0 & -- & 0.5 &  & \multicolumn{1}{l|}{\textit{reddit}} & 0.5917 & 0.7222 & 0.3659 & 0 & -- & 0.6667 \\
\textit{speech} & 0.6196 & 0.7727 & 0.4773 & 0.5 & -- & -- &  & \multicolumn{1}{l|}{\textit{speech}} & 0.6739 & 0.7727 & 0.5682 & 0.75 & -- & -- \\
\textit{textbook} & 0.6667 & 0.8070 & 0.4634 & 0.5 & -- & 1 &  & \multicolumn{1}{l|}{\textit{textbook}} & 0.5588 & 0.6491 & 0.4878 & 0 & -- & 0 \\
\textit{vlog} & 0.6486 & 0.6917 & 0.4615 & 0.5 & -- & -- &  & \multicolumn{1}{l|}{\textit{vlog}} & 0.5743 & 0.6417 & 0.3077 & 0 & -- & -- \\
\textit{voyage} & 0.6029 & 0.6667 & 0.5526 & -- & -- & -- &  & \multicolumn{1}{l|}{\textit{voyage}} & 0.6176 & 0.6 & 0.6316 & 0 & -- & 0 \\
\textit{how-to} & 0.7034 & 0.8281 & 0.5556 & -- & -- & -- & \multirow{-18}{*}{\textbf{}} & \multicolumn{1}{l|}{\textit{how-to}} & 0.6102 & 0.7188 & 0.4815 & -- & -- & -- \\ \bottomrule
\end{tabular}%
}
\caption{Accuracy of GDTB \texttt{test} by Genres and Relation Types. ``--'' means such relation types are not available.}
\label{tab:gdtb-scores}
\end{table*}

\end{document}